# Handling Imbalanced Dataset in Multi-label Text Categorization using Bagging and Adaptive Boosting

Genta Indra Winata
School of Electrical Engineering and Informatics
Institut Teknologi Bandung
Bandung, Indonesia
gentaindrawinata@gmail.com

Masayu Leylia Khodra
School of Electrical Engineering and Informatics
Institut Teknologi Bandung
Bandung, Indonesia
masayu@stei.itb.ac.id

*Abstract*— **Imbalanced dataset is occurred due to uneven distribution of data available in the real world such as disposition of complaints on government offices in Bandung. Consequently, multi-label text categorization algorithms may not produce the best performance because classifiers tend to be weighed down by the majority of the data and ignore the minority. In this paper, Bagging and Adaptive Boosting algorithms are employed to handle the issue and improve the performance of text categorization. The result is evaluated with four evaluation metrics such as hamming loss, subset accuracy, example-based accuracy and micro-averaged f-measure. Bagging.ML-LP with SMO weak classifier is the best performer in terms of subset accuracy and example-based accuracy. Bagging.ML-BR with SMO weak classifier has the best micro-averaged f-measure among all. In other hand, AdaBoost.MH with J48 weak classifier has the lowest hamming loss value. Thus, both algorithms have high potential in boosting the performance of text categorization, but only for certain weak classifiers. However, bagging has more potential than adaptive boosting in increasing the accuracy of minority labels.**

## I. Introduction

Text categorization has become one of the most decent method to organize text data since this can save time and has been proven in solving many complex problems. Nowadays, text data in several domains can be associated with a set of labels. The text categorization to set of labels are usually called multi-label categorization. For instance, a complaint text is categorized to some relevant government departments that are responsible in solving the complaint.

One of the most popular issue that catches the attention of academia and industry is imbalanced learning problem. This issue is one of the major problem because standard classifiers tend to be weighed down by majority classes and ignores the small ones [1]. In other words, the classifier will not give optimum expected accuracy to all classes. Most standard classifiers expect balanced class distribution and cannot handle the issue automatically.

There are several approaches of the current research activities such as data level approach and algorithm level approach [1]. The former concerns in balancing the class distribution through sampling and the latter approach modifies algorithm by adjusting weight or cost of various classes [2]. Algorithm level approach is better than data level approach in terms of avoiding *overfitting* issue [3]. *Overfitting* is a condition when model generalizes poorly to new data despite excellent performance over the training data [4]. This issue is commonly occurred due to *undersampling* by the duplication of the minority instances [5].

To our knowledge, there are only 2 previous researches [6], [7] which focus on handling imbalanced dataset for multi-label text categorization. This paper uses LAPOR dataset which was used before in previous research [8], but [8] did not handle the issue. On the other hand, we consider the dataset is too small to generalize all possible occurrences because [8] only has 2230 data. Thus, we acquired 2921 more instances from the institution who collects those data. The dataset consists of complaint text and set of government departments that are responsible in solving the complaint. The dataset consists of 70 different classes and 5151 instances. The distribution of the classes is imbalanced since there are 55 out of 70 minority classes and some of them only have 1 instance. The aim of our experiment is to evaluate the capability of Bagging and Adaptive Boosting algorithms in increasing the performance of multi-label text categorization.

In this paper, we employ algorithm level approach to LAPOR dataset limited to *Adaptive Boosting* and *Bagging*. The study evaluates the effectiveness of the both techniques and compares it with the baseline. For the baseline comparison, popular multi-label algorithms which include *Binary Relevance* and *Label Powerset* are used.

In the subsequent sections, we describe related works for handling multi-label text categorization and imbalanced datasets in section II. Then, we explain our method in section III. We analyze the characteristics of the dataset in section IV. In section V, we explain how we do the experiment. The result shows in section VI and at last, the conclusion is in section VII.

## II. RELATED WORK

Common approaches used in multi-label categorization are *algorithm adaptation* and *problem transformation* [9]. Algorithm adaptation is a method to modify existing single-label algorithm in order to be able to categorize multi-label datasets. ML-kNN is one example of algorithm adaptation. The algorithm splits the problem into independent single-label problems and each of them is classified using kNN algorithm. Problem transformation is a technique to split multi-label problem into one or more single-label problems. Some examples of problem transformation are *Label Powerset* and *Binary Relevance*. *Binary Relevance* (BR) maps multi-label problem into one or more independent binary single-label classification problem [10]. While *Label Powerset* (LP) is transforming multi-label problem into one multi-class labels present in the training instances [10].

To handle imbalanced dataset, there are several approaches such as data level approach and algorithm level approach [1]. The first approach concerns in create balance dataset distribution by sampling and the second approach focuses in modifying algorithm by adjusting weight and cost to handle imbalanced dataset [2]. There are some methods of sampling such as *undersampling* and *oversampling* [5]. *Oversampling* is a method to eliminate majority classes randomly to achieve equal distribution. Conversely, *undersampling* is a method to replicate minority classes to achieve equal distribution with the majority classes. Techniques that are classified as algorithm level approach are *adaptive boosting*, *bagging, cost-sensitive* and *active learning*. In this paper, we only consider *adaptive boosting* and *bagging* in our experiment and here are the explanation for each technique.

*Adaptive boosting* is an algorithm to find highly accurate classification rule by combining many weak hypotheses [6]. The training instances and their labels are weighted and if they are hard to be predicted correctly, they get higher weights and in opposition, if they are easy to be predicted correctly, they get lower weights. This process runs incrementally. One of the variant of boosting algorithm for multi-label classification problem is AdaBoost.MH [6]. The experiment results with the boosting algorithms dramatically reduces training error when a large amount of data is available and on the other hand, on small and noisy datasets, the rapid decrease of training error is often followed by *overfitting* [7]. In addition, Syaripudin and Khodra states adaptive boosting is able to handle imbalanced dataset, especially for single-label categorization [11].

*Bagging* is an algorithm for generating multiple bootstraps and use the aggregation average over all bootstraps to predict a class [12]. The hypothesis is obtained by voting. One of the variant of bagging algorithm for multi-label classification problem is Bagging.ML. [12] states *bagging* can significantly reduce error rate. Increasing number of iteration up to 25 bootstraps can reduce error rate, but more than that the error rate remains unchanging [12].

## III. OUR METHODS

We propose two techniques to solve imbalanced dataset issue: *Adaptive Boosting* and *Bagging*. Furthermore, each technique will be compared with the baseline algorithms which are state of the art for multi-label categorization such as *Binary Relevance* and *Label Powerset*. Overall, the research is conducted in several steps: data preprocessing, training, testing, and evaluation. Fig. 1 illustrates the flow of the entire experiment.

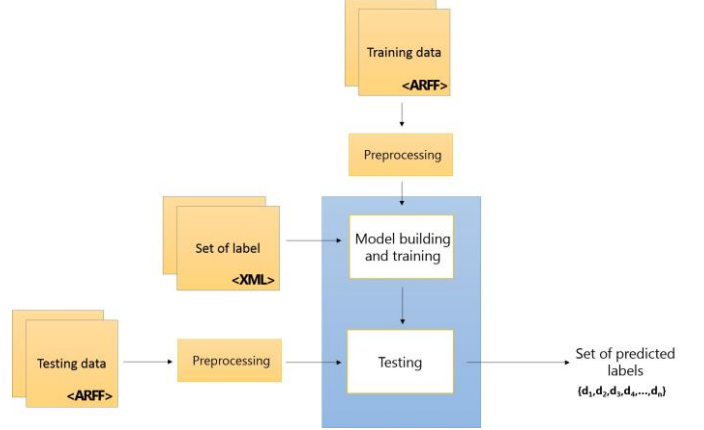

Fig. 1. Multi-label text categorization process

Data preprocessing covers 5 different processes such as sentence tokenization, text formalization, stopwords elimination, word stemming, and feature selection. Firstly, we extract token from complaint text. This process is conducted by splitting words from space character. Overall, we acquired 12982 different features from the text. Secondly, we formalize each word by using NLP-INA library. Thirdly, every stopwords in the complaint text is eliminated and then, we stem each word. Finally, from 12982 features, we select 753 features by selecting from its information gain value [13].

In next step, we train the model by using training dataset. The model is built using techniques which have been mentioned before. We test the model with the testing dataset. The result of this step is set of predicted labels. After we received predictions, we evaluated the model by using four metric evaluations such as hamming loss, subset accuracy, example-based accuracy and micro-averaged f-measure. Hamming loss and micro-averaged f-measure are used to measure the performance of imbalanced dataset approaches. Subset accuracy and example-based accuracy are used to determine the performance of the classification.

On the following equations we use following definitions. N is the number of instances, $h(x_i)$ is the set of predicted label, $y_i$ is the actual set of label and Q is the total number of possible class labels. *Hamming loss* evaluates number of times an example-label pair is misclassified [14]. (1) shows how *hamming loss* is calculated where ∆ stands for symmetric difference between 2 sets [14] or in other words, it is a xor operator.

$$hamming\_loss(h) = \frac{1}{N}\sum_{i=1}^{N}|h(x_i)\Delta y_i| \quad (1)$$

*Subset accuracy* evaluates how many instances are correctly classified [14].

$$subset\_accuracy(h) = \frac{1}{N}\sum_{i=1}^{N} I(h(x_i) = y_i) \qquad (2)$$

This metric evaluation is very strict as all labels must be correct classified. $I(h(x_i) = y_i) = 1$ if the predicted labels are equal matched with the actual labels, otherwise it is 0 [14].

Example-based accuracy evaluates number of labels which are correctly classified [14].

$$accuracy(h) = \frac{1}{N}\sum_{i=1}^{N} \frac{|h(x_i) \cap y_i|}{|h(x_i) \cup y_i|} \qquad (3)$$

Micro-averaged f-measure calculates the harmonic mean between *micro_precision* and *micro_recall* [14]. (4), (5) and (6) shows how to calculate the metric where $t_p$, $f_p$ and $f_n$ are the number of true positives, false positives and false negatives.

$$micro_{F_1} = \frac{2 \times micro_{precision} \times micro_{recall}}{micro_{precision} + micro_{recall}} \qquad (4)$$

$$micro_{precision} = \frac{\sum_{j=1}^{Q} tp_j}{\sum_{j=1}^{Q} tp_j + \sum_{j=1}^{Q} fp_j} \qquad (5)$$

$$micro_{recall} = \frac{\sum_{j=1}^{Q} tp_j}{\sum_{j=1}^{Q} tp_j + \sum_{j=1}^{Q} fn_j} \qquad (6)$$

## IV. LAPOR Dataset

We received LAPOR dataset from President's Delivery Unit for Development Monitoring and Oversight Indonesia [15]. The data was collected from October 2013 until December 2014. This dataset consists of 5151 *instances* with 70 classes. Each instance comprises features such as *id*, *complaint text*, complaint *topic* and *a set of label*. The dataset is imbalanced, since the distribution of classes is not equal and many minority classes are outnumbered to the majority classes. In order to tell this, we pick two different classes, "*Dinas Bina Marga dan Pengairan (DBMP) Kota Bandung*" class which has 1304 instances and "*Bagian Pembangunan dan Sumber Daya Alam (BagPem) Kota Bandung*" class which has only 1 instance. Fig. 2 shows sample data. The sample data can be categorized into 2 class labels: "*Dinas Sosial (Dinsos) Kota Bandung*" and "*Satuan Polisi Pamong Praja (Satpol PP) Kota Bandung*".

> *Di stasiun KA Kiaracondong (jln Kiaracondong antara kebaktian s d kantor Polisi kebon jayanti) ada gepeng anak usia sekolah ngelem, bicaranya kasar tdk karuan, suka ganggu penumpang wanita! Trm ksh!*

Fig. 2. Sample data

On average, class labels have 103.3 instances and there are 15 out of 70 labels which have higher number of instances than average. In other words, the dataset has lack of density characteristic. The lack of density problem happened when the induction algorithms do not have enough data to make generalizations about the distribution of instances. This issue is occurred when the concentration of minority instances within the dataset are so low compared to majority instances [16]. The threat of this issue is the minority instances are tend to be treated as noise by classifier during the categorization process. Fig. 3 illustrates the distribution of label within the dataset and the horizontal axis shows the labels sorted in descending order from label with the highest to the lowest number of instances.

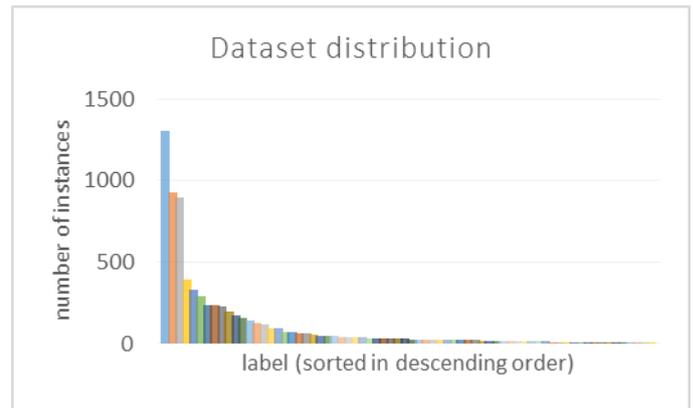

Fig. 3. Dataset distribution

Most of the complaints use Indonesian language. Informal words, abbreviations, and stopwords are frequently found inside the text. Therefore, preprocess is required. We apply binary term weighting and features are selected based on information gain calculation for each term.

## V. Experiment

Our experiment used dataset divided into training dataset and testing dataset. Training dataset and testing dataset are consists of 4120 instances and 1031 instances. Fig. 4 and Fig. 5 illustrates the distribution of label within the training data and test data respectively and the horizontal axis shows the labels sorted in descending order from label with the highest to the lowest number of instances.

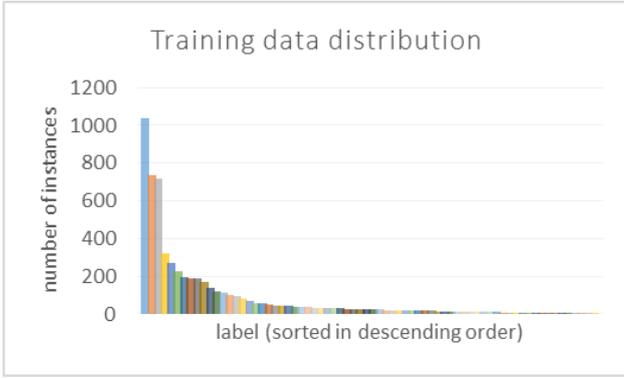

Fig. 4. Training data distribution

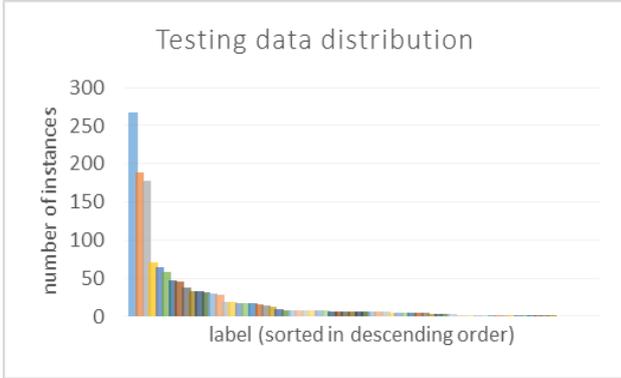

Fig. 5. Testing data distribution

The experiment scenario uses five weak classifiers: Decision Stump [17], J48 [17], Random Forest [18], Naive Bayes [19] and SMO [20]. In each weak classifier we applies them to multi-label classifiers such as *Binary Relevance* and *Label Powerset* [10]. We use two respective algorithms as our baseline and are compared with two imbalanced dataset handling approaches: *adaptive boosting* and *bagging*. We selected AdaBoost.MH [6] and Bagging.ML [12] as the approaches algorithm. For each approach, we test up to 20 iterations iteratively and measure its performance. We use MULAN [21] to employ AdaBoost.MH and MEKA library [22] to employ Bagging.ML in our experiment.

## VI. RESULT & DISCUSSION

We obtain four metric evaluations from experiment. The result are shown below. *Binary Relevance* and *Label Powerset* are denoted as BR and LP respectively.

TABLE I. HAMMING LOSS

| Weak classifier | Baseline BR | Baseline LP | AdaBoost.MH | Bagging.ML (BR) | Bagging.ML (LP) |
|---|---|---|---|---|---|
| Decision Stump | 0.0152 | 0.0247 | 0.0197 | N/A | 0.0250 |
| J48 | 0.0133 | 0.0188 | 0.0131 | 0.0150 | 0.0170 |
| Random Forest | 0.0132 | 0.0179 | 0.0146 | N/A | N/A |
| Naive Bayes | 0.0420 | 0.0159 | 0.0327 | N/A | N/A |
| SMO | 0.0144 | 0.0148 | 0.0197 | 0.0150 | 0.0150 |

NOTE — N/A (not applicable), the particular classifier was not used and shaded cell represents the classifier which outperformed the baseline.

TABLE II. SUBSET ACCURACY

| Weak classifier | Baseline BR | Baseline LP | AdaBoost.MH | Bagging.ML (BR) | Bagging.ML (LP) |
|---|---|---|---|---|---|
| Decision Stump | 0.3346 | 0.2318 | 0.0039 | N/A | 0.2320 |
| J48 | 0.4074 | 0.4016 | 0.4277 | 0.3750 | 0.3800 |
| Random Forest | 0.4103 | 0.3986 | 0.3337 | N/A | N/A |
| Naive Bayes | 0.2144 | 0.4219 | 0.0145 | N/A | N/A |
| SMO | 0.4200 | 0.4588 | 0.0039 | 0.4000 | 0.4490 |

NOTE — N/A (not applicable), the particular classifier was not used and shaded cell represents the classifier which outperformed the baseline.

TABLE III. EXAMPLE-BASED ACCURACY

| Weak classifier | Baseline BR | Baseline LP | AdaBoost.MH | Bagging.ML (BR) | Bagging.ML (LP) |
|---|---|---|---|---|---|
| Decision Stump | 0.4123 | 0.2726 | 0.0039 | N/A | 0.2730 |
| J48 | 0.5151 | 0.5034 | 0.5301 | 0.5520 | 0.5400 |
| Random Forest | 0.5080 | 0.4907 | 0.3847 | N/A | N/A |
| Naive Bayes | 0.4098 | 0.5570 | 0.0417 | N/A | N/A |
| SMO | 0.5556 | 0.5821 | 0.0039 | 0.5740 | 0.5850 |

NOTE — N/A (not applicable), the particular classifier was not used and shaded cell represents the classifier which outperformed the baseline.

TABLE IV. MICRO-AVERAGED F-MEASURE

| Weak classifier | Baseline BR | Baseline LP | AdaBoost.MH | Bagging.ML (BR) | Bagging.ML (LP) |
|---|---|---|---|---|---|
| Decision Stump | 0.4919 | 0.2716 | 0 | N/A | 0.2720 |
| J48 | 0.5950 | 0.5055 | 0.6044 | 0.6240 | 0.5700 |
| Random Forest | 0.5839 | 0.4988 | 0.4704 | N/A | N/A |
| Naive Bayes | 0.3889 | 0.5801 | 0.1094 | N/A | N/A |
| SMO | 0.6095 | 0.5977 | 0 | 0.6270 | 0.6040 |

NOTE — N/A (not applicable), the particular classifier was not used and shaded cell represents the classifier which outperformed the baseline.

In general, the results obtained using imbalanced dataset handling approach using adaptive boosting improves the performance using J48 algorithm, but this approach did not work the same for other classifiers. Hamming loss value of J48 is slightly reduced for 0.0002, subset accuracy, example-based accuracy and micro-averaged f-measure value are raised for 0.0203, 0.0150 and 0.0094 respectively compared to BR.

There was an increase in almost all metric measurements, especially a reduction in hamming loss value and increase subset accuracy and example-based accuracy on adaptive boosting approach to J48 algorithm. Bagging algorithm increases the micro-averaged f-measure at J48 algorithm with very significant (See Table I). Adaptive boosting method has poor performance in the SMO algorithm, while bagging method can improve the accuracy and example-based micro-averaged f-measure SMO algorithms. Generally, the performance of weak classifiers such as *decision stump*, *naive bayes* and SMO are weaken after using *adaptive boosting* and *bagging*.

Overall, SMO with Bagging.ML-LP has the best performance in terms of accuracy subset, example-based accuracy and micro-average f-measure (See Table II, Table III and Table IV). In other hand, J48 algorithm with

AdaBoost.MH has the lowest hamming loss value (See Table I).

Fig. 6 and Fig. 7 shows the performance of categorization trend for J48 algorithm using baseline and imbalanced dataset handling approaches to the number of iterations. We only show J48 result since this algorithm gives the highest performance improvements among all.

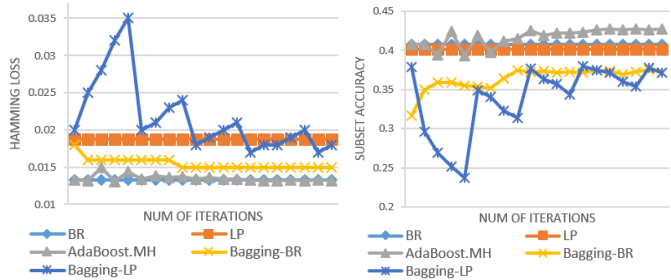

Fig. 6. J48 hamming loss (left) and subset accuracy (right)

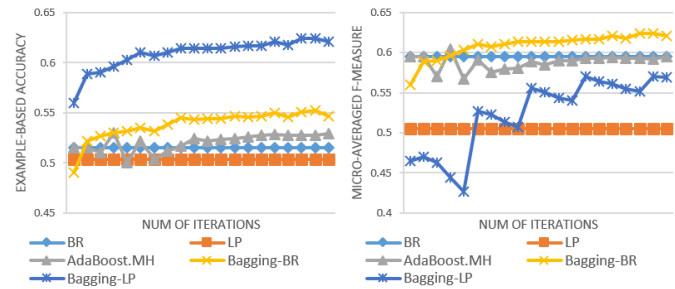

Fig. 7. J48 example-based accuracy (left) and micro-averaged f-measure (right)

The experiment results for J48 algorithm can be seen in Fig. 4 The results demonstrates Bagging.ML algorithm fluctuates and is getting closer to an optimum. Additionally, AdaBoost.MH algorithm is better than the baseline algorithms. On the other hand, Bagging.ML algorithm has better performance than the baseline for all metrics except subset accuracy.

Fig. 8 illustrates the accuracy margin between J48 with BR and AdaBoost.MH for each label. The labels are sorted in ascending order from label with the lowest to the highest number of instances. Some of the majority labels have significant raise of accuracy. Conversely, there are only small number of labels which have accuracy drop. The accuracy drop is not significant because the accumulation of changes is positive. The opposite condition was occurred for the minority labels. There is only one minority label which has positive change.

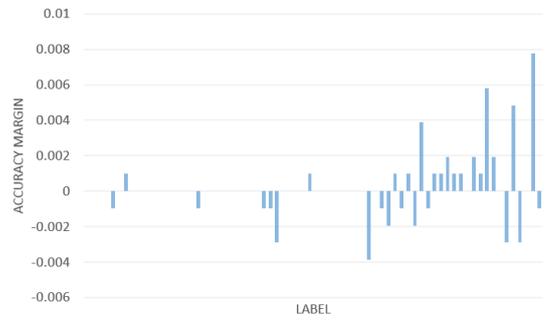

Fig. 8. The accuracy margin between BR and AdaBoost.MH with J48 algorithm

Fig. 9 illustrates the accuracy margin between SMO with LP and Bagging-LP for each label. The labels are sorted in ascending order from label with the lowest to the highest number of instances. Most of minority labels have slight changes and there are only small number of instances which have huge ups and downs. Most of the changes are within the majority labels. There are only very few labels which have lower accuracy than baseline. Overall, the accuracy of the labels are raised, especially for majority labels. From the above explanation, we could take conclusion that adaptive boosting is less potential in increasing the accuracy of minority than bagging.

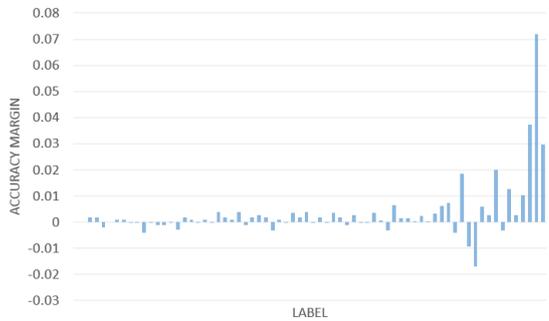

Fig. 9. The accuracy margin between LP and Bagging-LP with SMO algorithm

## VII. CONCLUSION

In this paper, two imbalanced dataset handling approaches such as bagging and adaptive boosting are evaluated and compared with multi-label classifiers. As discussed in Section VI, both approaches can improve the performance of the categorizations, especially for J48 and SMO. Overall, SMO with Bagging.ML-LP has the best performance in terms of subset accuracy and example-based accuracy. SMO with Bagging.ML-BR has the best micro-averaged f-measure value among all. In other hand, J48 algorithm with AdaBoost.MH has the lowest hamming loss value.

In J48 case, adaptive boosting technique can impressively boost the accuracy of most majority labels, but in the same time the accuracy of the majority labels are reduced. The reduction in accuracy is not significant since it is still overpowered by the increment of accuracy. There is only one label which has positive change and still it is insignificant. In other hand, bagging algorithm increases the accuracy of most labels significantly and this is only occurred for SMO classifier. For other weak classifiers, bagging tends to have worse performance than the baseline. At last, we can conclude that bagging has more potential than adaptive boosting in increasing the accuracy of minority labels.